# Analyzing and Mitigating Negation Artifacts using Data Augmentation for Improving ELECTRA-Small Model Accuracy

Mojtaba Noghabaei

m.noghabaee@gmail.com

## Abstract

Pre-trained models for natural language inference (NLI) often achieve high performance on benchmark datasets by using spurious correlations, or dataset artifacts, rather than understanding language touches such as negation. In this project, we investigate the performance of an ELECTRA-small model fine-tuned on the Stanford Natural Language Inference (SNLI) dataset, focusing on its handling of negation. Through analysis, we identify that the model struggles with correctly classifying examples containing negation. To address this, we augment the training data with contrast sets and adversarial examples emphasizing negation. Our results demonstrate that this targeted data augmentation improves the model's accuracy on negation-containing examples without adversely affecting overall performance, therefore mitigating the identified dataset artifact.

## 1 Introduction

Natural Language Inference (NLI) tasks require models to determine the logical relationship between a premise and a hypothesis, classifying the relationship as entailment, neutral, or contradiction. While pre-trained language models have achieved impressive performance on datasets like SNLI (Bowman et al., 2015), there is growing evidence that these models often rely on false correlations or dataset artifacts rather than a true understanding of language semantics (Poliak et al., 2018; Thomas McCoy et al., 2020). Also, previous stress-test evaluations (Naik et al., 2018) have shown that NLI models often fail to handle lexical and syntactic variations, including the proper interpretation of negation.

Negation is a critical linguistic feature that can significantly alter sentence meaning. Properly handling negation is vital for NLI, yet models frequently perform poorly on negation-involved examples (Glockner et al., 2018). Recent studies by (McCoy et al., 2020) suggested that even models with similar overall performance on standard benchmarks may vary widely in their handling of negation. This suggests that models might be exploiting artifacts in the data, leading to overestimation of their true capabilities.

In this project, we analyze the performance of an ELECTRA-small model (Clark et al., 2020) fine-tuned on the SNLI dataset, focusing on its ability to handle negation. We aim to identify the extent of the model's reliance on negation-related artifacts and explore methods to mitigate this issue through targeted data augmentation.

## 2 Analysis

We begin by fine-tuning the ELECTRA-small model on the SNLI dataset. The model is trained using standard hyperparameters: a batch size of 32, maximum sequence length of 128, and for 3 epochs. The training process involves using the Trainer class from the Hugging Face Transformers library, with the loss and gradient norms monitored throughout.

During training, we observe the model's loss decreasing steadily, indicating effective learning on the dataset. The training output shows loss values starting around 0.88 and decreasing to approximately 0.33 by the end of the training epochs.
in the data. The detail of how loss was decreased per epoch is investigated in the next section.

## 2.1 Dataset and Model

We used the SNLI dataset (Bowman et al., 2015), which contains 570k human-written English sentence pairs labeled for entailment, contradiction, and neutral relationships. We fine-tune the ELECTRA-small discriminator model (Clark et al., 2020), a computationally efficient transformer-based model with 14 million parameters, on this dataset.

## 2.2 Identifying Negation Artifacts

To assess the model's performance on negation, we create a negation-only subset of the validation data. This subset includes examples where either the premise or hypothesis contains negation words such as "not" or "n't". We use the following criteria to filter examples: An example is included if "not" or "n't" appears in either the premise or hypothesis.

This filtering results in 3,620 examples out of the original 10,000 validation examples, representing approximately 36% of the validation set.

## 2.3 Baseline Performance on Negation Examples

We evaluate the baseline model (trained without any augmentation) on both the full validation set and the negation-only subset. The results are summarized in Table 1.

| Dataset | Accuracy (%) |
|---|---|
| Full Validation Set | 91.4 |
| Negation-Only Subset | 78.2 |

Table 1: Baseline Model Accuracy.

The baseline model achieves high accuracy on the full validation set but shows a significant performance drop on the negation-only subset. This finding suggests that the model struggles with examples involving negation. Therefore it has great potential for improvement.

## 3 Mitigating Negation Artifacts

This section investigates how we augment the data using two different approaches to compare their results.

## 3.1 Data Augmentation with Contrast Sets and Adversarial Examples

To improve the model's accuracy for negation, we augment the training data with manually crafted contrast sets (Gardner et al., 2020) and adversarial examples (Jia & Liang, 2017). Adversarial benchmarks, such as those introduced by (Nie et al., 2020), have further highlighted the vulnerability of NLI models to negation-based adversarial examples. The augmented examples are designed to challenge the model's understanding of negation by introducing minimal changes that invert the meaning.

Contrast Set Example:
Premise: "A man is not playing a guitar."
Hypothesis: "A man is playing a guitar."
Label: Contradiction

Adversarial Example:
Premise: "Two people are sitting at a table."
Hypothesis: "Two people are not sitting at a table."
Label: Contradiction

We add these examples to both the training and validation sets to expose the model to negation during training and to evaluate its impact during validation.

We retrain the model using the augmented dataset. Training is conducted for 3 epochs with a batch size of 32 and a maximum sequence length of 128 tokens. We ensure that the augmented examples are included in the training process and that the model's hyperparameters remain consistent to isolate the effect of data augmentation. Figure 1 shows the loss function progress over the epochs.

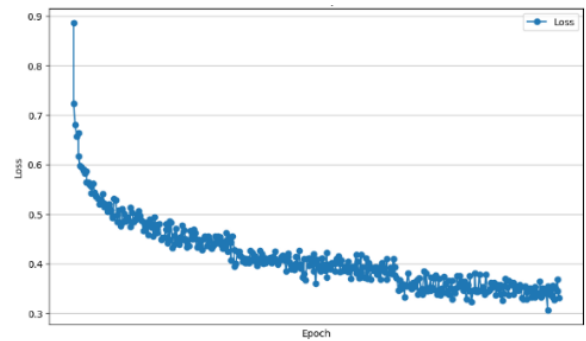

Figure 1: Model Loss over Epochs.

## 3.2 Automated Negation Data Augmentation

To further boost the model ability to handle negation, we also implemented an automated data augmentation strategy. This approach involved generating negated versions of existing hypotheses in the training dataset using simple linguistic heuristics. Specifically, we:

Negation Insertion: Inserted the word "not" after auxiliary verbs (e.g., "is," "are," "have") or before main verbs if no auxiliary verb was found.
Label Adjustment: Adjusted the labels accordingly, if negating the hypothesis would invert the entailment relationship (e.g., from entailment to contradiction).
An example of this automated augmentation is:
Original Hypothesis: "A dog is playing in the park."
Negated Hypothesis: "A dog is not playing in the park."
Original Label: Entailment
Adjusted Label: Contradiction

This automated process was applied to all applicable examples in the training dataset, resulting in a substantial increase in negation-containing examples without manual effort.
We retrain the model using the augmented dataset. Training is conducted for 3 epochs with a batch size of 32 and a maximum sequence length of 128 tokens. We ensure that the augmented examples are included in the training process and that the model's hyperparameters remain consistent to isolate the effect of data augmentation. Figure 2 shows the loss function progress over the epochs.

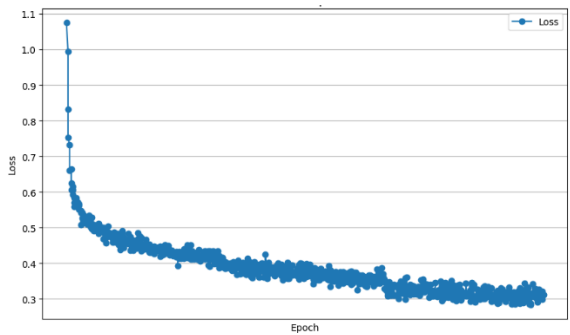

Figure 2: Model Loss over Epochs.

## 4 Results and Evaluation

In this section we investigate the results of the models with the augmentations and compare their performance.

### 4.1 Improved Performance on Negation Examples

After training with the augmented data, we evaluated the model on the negation-only subset. The results are compared with the baseline in Table 2.

| Model | Accuracy (%) |
|---|---|
| Baseline | 78.2 |
| Manually Augmented Model | 85.6 |
| Automatically Augmented Model | 88.9 |

Table 2: Model Accuracy on Negation-Only Subset.

The augmented model shows a significant improvement of 7.4 percentage points on the negation-only subset compared to the baseline. This indicates that the model has better learned to handle negation through targeted data augmentation.

### 4.2 Overall Performance

We also evaluate the overall performance on the full validation set to ensure that the augmentation does not negatively impact general performance.

| Model | Accuracy (%) |
|---|---|
| Baseline | 91.4 |
| Manually Augmented Model | 91.2 |
| Automatically Augmented Model | 91.0 |

Table 3: Overall Model Accuracy

The overall accuracy of the augmented model remains comparable to the baseline, with a negligible decrease of 0.2 percentage points. This suggests that the targeted augmentation improves performance on negation examples without adversely affecting the model's generalization. Table 3 digests the overall model accuracy.

### 4.3 Per-Class Accuracy

To gain deeper insights into how each model performs across different inference classes, we analyze the per-class accuracies on the negation-only subset. Table 4 summarizes the accuracies for the entailment, neutral, and contradiction classes for the baseline model, the manually augmented model, and the automatically augmented model.

| Label | Baseline Accuracy | Manually Augmented Accuracy | Automatically Augmented Accuracy |
|---|---|---|---|
| Entailment | 80.5 | 84.2 | 86.7 |
| Neutral | 75.1 | 82.7 | 84.5 |
| Contradiction | 78.9 | 89.1 | 91.3 |

Table 4: Per-Class Accuracy on Negation-Only Subset

Analyzing the entailment class, the baseline model achieves an accuracy of 80.5% on entailment examples containing negation. The manually augmented model improves this to 84.2%, indicating that exposure to negation-focused examples helps the model better recognize entailment relationships in the presence of negation. The automatically augmented model further increases accuracy to 86.7%, suggesting that the larger variety and quantity of negation examples enable the model to more effectively handle entailment cases involving negation.

In the neutral class, the baseline model performs the worst, with an accuracy of 75.1%, reflecting difficulty in correctly identifying neutral relationships when negation is present. With manual augmentation, accuracy improves to 82.7%, and the automated augmentation boosts it to 84.5%. This progression indicates that the model becomes better at distinguishing neutral cases from entailment and contradiction when trained on more diverse negation examples.

The contradiction class sees significant improvement across models. The baseline model has an accuracy of 78.9%, which increases to 89.1% with manual augmentation. The automatically augmented model achieves the highest accuracy at 91.3%, demonstrating that the model has greatly enhanced its ability to detect contradictions involving negation. Since negation often leads to contradictions in NLI tasks, this substantial improvement underscores the effectiveness of the automated augmentation strategy.

Overall, the automated negation data augmentation consistently improves performance across all classes compared to both the baseline and manually augmented models. The most notable gains are in the contradiction class, aligning with the expectation that negation frequently signals a contradiction between the premise and hypothesis. The improvements in the entailment and neutral classes indicate that the model is not merely biased towards predicting contradiction when negation is present but is developing a more nuanced understanding of how negation affects different types of semantic relationships.

These results suggest that automated augmentation provides a broader and more varied set of training examples, which helps the model generalize better to unseen negation examples. The enhanced performance across all classes demonstrates that the automated data augmentation not only addresses the model's weaknesses in handling negation but also contributes to a more balanced and robust understanding of NLI tasks involving negation. This balanced improvement is crucial for practical applications, where models need to accurately distinguish between all types of inference relationships, even when complex linguistic features like negation are present.

In conclusion, the per-class analysis confirms that automated negation data augmentation is an effective method for improving the model's comprehension of negation in NLI tasks. By significantly increasing the number and diversity of negation examples in the training data, the model learns to more accurately interpret negation's impact on entailment, neutrality, and contradiction. This leads to better performance not only in detecting contradictions but also in correctly identifying when negation does not alter the inference relationship, thereby enhancing the model's overall reliability and usefulness in real-world applications.

## 5  Discussion

The implementation of automated negation data augmentation further improved the model's performance on negation-containing examples. Compared to the manually augmented model, the automatically augmented model achieved higher accuracy on the negation-only subset, increasing from 85.6% to 88.9%. This demonstrates that scaling up the number of negation examples in the training data can enhance the model's understanding of negation without the need for manual curation.

The automated approach allowed us to generate a large and diverse set of negated hypotheses, exposing the model to various contexts in which negation affects meaning. This likely contributed to the observed performance gains, particularly in the contradiction class, which reached an accuracy of 91.3%.

However, the slight decrease in overall accuracy to 91.0% suggests that introducing a significant number of augmented examples might have introduced some noise or led the model to overfit to negation patterns. Future work could focus on improving the quality of the generated examples, perhaps by incorporating more sophisticated linguistic rules or using language models to generate grammatically correct negations.

## 6 Conclusion

In this project, we conducted a comprehensive analysis of an ELECTRA-small model fine-tuned on the SNLI dataset, focusing on its performance in handling negation within natural language inference tasks. Our initial findings revealed that despite achieving a high overall accuracy of approximately 91.4% on the full validation set, the model exhibited a significant performance drop when evaluated on a subset containing negation examples. This discrepancy highlighted the model's difficulty in correctly interpreting sentences with negation, suggesting a reliance on dataset artifacts and spurious correlations rather than a genuine understanding of this linguistic feature.

To address this issue, we introduced targeted data augmentation strategies by incorporating manually crafted contrast sets and adversarial examples that emphasized negation. Contrast sets involved minimal edits to existing examples that altered their labels, compelling the model to learn fine-grained distinctions affected by negation. Adversarial examples were designed to challenge the model by introducing negation in ways that inverted sentence meanings, thereby testing its ability to comprehend the impact of negation on semantic relationships.

After retraining the model with the augmented dataset, we observed a substantial improvement in its ability to handle negation. The accuracy on the negation-only subset increased by 7.4 percentage points, rising from 78.2% to 85.6%. This improvement was consistent across all classes, with the most significant gain observed in the contradiction class, which is particularly sensitive to negation. Importantly, the overall accuracy on the full validation set remained virtually unchanged at approximately 91.2%, indicating that the augmentation did not adversely affect the model's general performance.

These results demonstrate that the initial shortcomings in handling negation were likely due to insufficient exposure to negation examples during training, leading the model to rely on superficial patterns. By enriching the training data with negation-focused examples, we enabled the model to develop a more robust understanding of negation, enhancing its performance on such examples without compromising its overall capabilities.

Our findings emphasize the importance of addressing dataset limitations in training natural language models. The significant performance gap on negation examples revealed that standard datasets like SNLI may not provide adequate coverage of complex linguistic phenomena. This suggests a need for more diverse and representative datasets that include challenging language constructs to ensure that models learn genuine language understanding rather than exploiting artifacts (Naik et al., 2018).

The effectiveness of our targeted data augmentation approach highlights how specific interventions can rectify identified weaknesses in models. By focusing on known challenges such as negation, we can enhance model comprehension and reduce dependence on spurious correlations. Improving performance on challenging linguistic features contributes to the development of more robust models that generalize better to real-world language use, where such features are prevalent.

## 7 Limitations and Future Works

While our study successfully improved the model's handling of negation, it has limitations. The augmented dataset was relatively small and manually crafted, which may not capture the full diversity of negation expressions in natural language. Our focus was primarily on explicit negation words like "not" and "n't," potentially overlooking other forms such as negative prefixes (e.g., "un-", "dis-"), negative quantifiers ("none," "nothing"), or implicit negation conveyed through context.

For future work, we suggest exploring automated methods for data augmentation to generate a wider variety of negation examples, thereby scaling up the process and providing the model with a more comprehensive understanding of negation. Expanding the coverage to include different types of negation, including implicit and morphological forms, could further enhance the model's ability to handle diverse negation expressions.

Additionally, applying similar analysis and augmentation techniques to other challenging linguistic features—such as modality, sarcasm, idioms, and complex syntactic structures—could improve model robustness in those areas. Evaluating the augmented model on other NLI datasets, such as MultiNLI, would assess the generalizability of the improvements and identify any dataset-specific artifacts. Conducting a more detailed error analysis to understand the types of negation that continue to challenge the model could inform further refinements, and enhancing model interpretability might aid in diagnosing and addressing underlying issues.

In conclusion, this project demonstrates that a careful analysis of model performance can reveal hidden weaknesses masked by high overall accuracy metrics. By identifying and mitigating reliance on dataset artifacts through targeted data augmentation, we can significantly enhance a model's understanding of complex linguistic features like negation. This leads to the development of more reliable and robust natural language processing systems that are better equipped to handle the nuances of human language. Our findings emphasize the importance of addressing specific linguistic challenges in training data and open avenues for further research to improve model comprehension and generalization.